\documentclass[conference]{IEEEtran}
\ifCLASSINFOpdf
  \usepackage[pdftex]{graphicx}
\else
\fi
\usepackage[nocompress]{cite}

\usepackage{amsmath}
\usepackage{amsfonts}
\usepackage{mathtools,xparse}
\usepackage[section]{placeins}

\DeclarePairedDelimiter{\norm}{\lVert}{\rVert}
\NewDocumentCommand{\normL}{ s O{} m }{%
  \IfBooleanTF{#1}{\norm*{#3}}{\norm[#2]{#3}}_{L_2(\Omega)}%
}

\begin{document}
%
\title{Towards Ophthalmologist Level Accurate Deep Learning
System for OCT Screening and Diagnosis}

\author{\IEEEauthorblockN{Mrinal Haloi}
\IEEEauthorblockA{CTO, Artelus\\
\{h.mrinal@iitg.ernet.in, mrinalhaloi@artelus.com\}}}


%


\maketitle

\begin{abstract}
In this work, we propose an advanced AI-based grading system for OCT images. The proposed system is a very deep fully convolutional attentive classification network trained with end-to-end advanced transfer learning with online random augmentation. It uses quasi-random augmentation that outputs confidence values
for diseases prevalence during inference. It’s a fully automated retinal OCT analysis AI system capable of pathological lesions understanding without any offline preprocessing/postprocessing step or manual feature extraction. We present a state-of-the-art performance on the publicly available Mendeley OCT dataset.
\end{abstract}


%
\IEEEpeerreviewmaketitle

\section{Introduction}
Sight-threatening retinal diseases are one of the major prevailed diseases among the population of varied age groups. Prevalence of retinal diseases among age groups and demographics are the main factor of vision loss.  Starting with diabetic retinopathy (DR), glaucoma, age-related macular degeneration (AMD) etc are the most common retinal diseases observed in society. Almost all of the retinal diseases can be identified using Optical coherence tomography (OCT) \cite{oct} scans. OCT is a non‐invasive optical imaging technique that is the optical analog of ultrasound imaging. This device provides high‐resolution cross‐sectional images of the retina, optic nerve head and retinal nerve fiber layer thickness (RNFL) that can be qualitatively and quantitatively evaluated.
It was originally developed to provide objective and quantitative estimates of the thickness of the RNFL. OCT RNFL measurements are reproducible and have been shown in cross-sectional studies to be able to discriminate glaucomatous from healthy eyes \cite{oct_review}.
With OCT, an ophthalmologist can see each of the distinctive layers of Retina.  This allows the ophthalmologist to map and measure their thickness. These measurements help with diagnosis and treatment guidance for glaucoma and other diseases of the retina. These retinal diseases include CNV, DME, and Drusen. 
Choroidal neovascularization (CNV) \cite {cnv} involves the growth of new blood vessels that originate from the choroid through a break in the Bruch membrane into the sub-retinal pigment epithelium (sub-RPE) or sub-retinal space. CNV is a major cause of visual loss.
Diabetic Macular Edema (DME) \cite{dme} occurs when fluid and protein deposits collect on or under the macula of the eye (a yellow central area of the retina) and cause it to thicken and swell (edema). The swelling may distort a person's central vision because the macula holds tightly packed cones that provide sharp, clear, central vision to enable a person to see detail, form, and color that is directly in the center of the field of view.
Drusen are yellow deposits under the retina. Drusen are made up of lipids, a fatty protein. Drusen likely do not cause AMD.  But having drusen increases a person’s risk \cite{drusen} of developing AMD.
There are different kinds of drusen. “Hard” drusen are small, distinct and far away from one another. This type of drusen may not cause vision problems for a long time, if at all.
“Soft” drusen are large and cluster closer together. Their edges are not as clearly defined as hard drusen. This soft type of drusen increases the risk of AMD.
OCT also represents a promising new technology for imaging vascular microstructure \cite{oct_bio} with a level of resolution not previously achieved with the use of other imaging modalities.
Considering the usefulness of OCT in detecting almost all retinal diseases we propose to use AI in the OCT analysis domain. We have developed an AI model that can be used to detect the presence of CNV, DME and Drusen can help doctors in diagnosing giving correct treatment, which can be used as a screening tool in clinics making adequate healthcare available to all the population and reducing the burden on doctors.
The proposed model builds upon recent advancement of deep learning and computer vision. Our novel deep model can outperform human ophthalmologist in the field of oct analysis.

\begin{figure}[h]
  \centering
      \includegraphics[width=3.6in,height=1.6in]{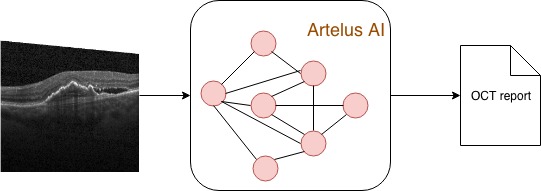}
\caption{Artelus AI OCT screening}
\label{fig:artelusoct}
\end{figure}

\subsection{Contribution}
\textbf{OCT classification learning:} We present an ophthalmologist performance level OCT AIAS system for eye diseases diagnosis.
  Following are the advantages of the proposed AIAS system?
    \begin{itemize}
    \item Trained on a small number of samples with better generalization performance.
    \item A novel deep learning algorithm that outperform average human ophthalmologists.
    \item Faster and deployable in a mobile chip.
    \item Accessible healthcare in remote areas. 
    \end{itemize}

\subsection{Related Work}
Deep learning is successfully applied in many applications of the medical image analysis domain. Tasks related to features segmentation \cite{retina_work} and feature classification \cite{retina_work4} can be performed using both traditional computer vision and deep learning. Deep learning recently captured momentum to outperform traditional features based methods used in the medical image domain. Starting from retinal analysis to cancer analysis, deep learning proved its usefulness as screening and diagnostic tools for doctors. In the eye \cite{retina_work2} and X-ray \cite{xray_work} screening scenarios deep learning outperforms human practitioners. 

Deep learning is also widely used in OCT image analysis task. A deep learning based OCT image segmentation method \cite{retina_work3} proposed pathological lesions segmentation using a fully convolutional method. Intraretinal cystoid fluid and subretinal fluid detection deep learning method \cite{retina_work4} using OCT was successfully tested with high accuracy. 
In case patient referral recommendation for sight-threatening retinal diseases using OCT images \cite{retina_work5}, this deep learning method outperforms expert ophthalmologist.

\section{Method}

\subsection{Dataset}

\begin{itemize}
\item Mendeley dataset \cite{mendeley}: It has total 84495 OCT images with 3 diseases and 1 normal class.To train the model 75K of these images were used. Extensive data augmentation techniques with more than 230 convolutional layers were used to train the deep model for CNV, DME, DRUSEN and Normal OCT image classification.
A validation set of total 8346 images with 2631 normal images and 5715 images with CNV, DME, and Drusen was used to check the accuracy of the model in both binary classification and multiclass classification. 
A test set of total 1000 images with 250 normal images and 750 images with CNV, DME, and DRUSEN was used to test the accuracy of the model in both binary classification and multiclass classification.
\end{itemize}

\subsection{Proposed Algorithm} Deep learning is proved to be the current state-of-the-art for
computer vision/image processing, speech, text problems and automotive. We
propose a very deep network that has around 50
million trainable parameters and 250 layers deep with custom convolutional blocks. The proposed
network is a modified version of network architecture presented in \cite{haloi}. Values of the network parameters are learned using Adam optimizer algorithm. The neural network hyper-parameters learning is an iterative process based on loss value computed between ground truths given by ophthalmologists and network predictions. This loss value acts as a feedback to the optimizer to learn most representative network parameters to understand retinal pathological lesions.

\subsection{Network Architecture}
\label{net_arch}
\textbf{Encoder Architecture for Transfer Learning} One of the concerns with medical data is the unavailability of high volume high quality labeled data. Even the available label by one single ophthalmologist might not reflect the same opinion as by another fellow practitioner. For an effective system, we need data to be labeled by many ophthalmologists. Since it's very expensive to get high volume data annotated by multiple annotators, we have another effective approach to deal with the same; label noise minimization. We label a fraction of the dataset by multiple annotators (Golden set) and leave out the rest of the dataset with single annotation (TFL set). We train a two heads model with one head to reproduce the sample using a variational encoder and the second head to classify the sample on the whole dataset. Sample-wise weighted loss is used for the classification head. We set high loss weight for Golden set samples and relatively low loss weight for TFL set samples. The weighting strategy is based on sample annotation. The trained encoder parameters will be used for the initialization of the final deep network for classification. This way we make sure that all samples pathological features information are preserved in the encoder network and can be transferred to the final network.

\begin{figure}[h]
  \centering
      \includegraphics[width=3.6in,height=3.6in]{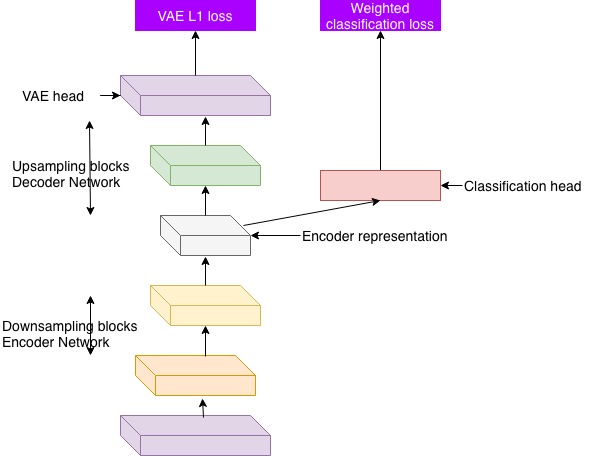}
\caption{Encoder based transfer learning architecture.}
\label{fig:artelusoct2}
\end{figure}

\begin{figure}[h]
  \centering
      \includegraphics[width=3.1in,height=2.9in]{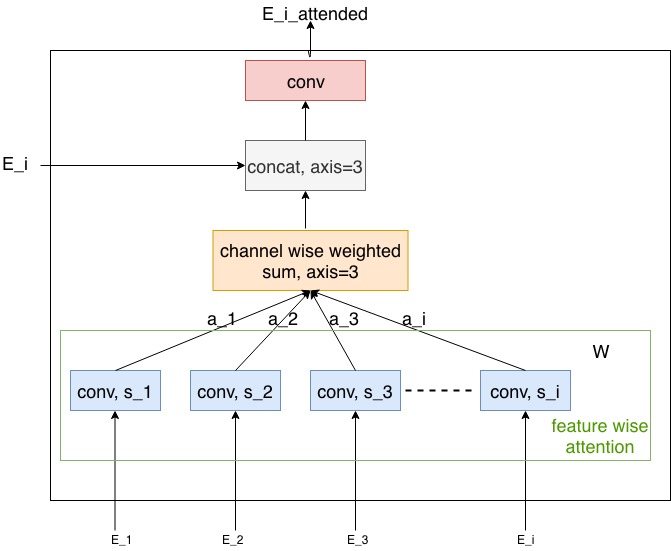}
\caption{Partial attention over feature maps}
\label{fig:artelus_oct3}
\end{figure}

\textbf{Partial attention}
We use a modified version of the partial attention \cite{partial_attention} over the lower blocks of the classifier feature maps. Feeding the intermediate feature maps information to the higher layers improve the learning stability and the representation power of the network. It also facilitates easier training of very deep model by eliminating the vanishing gradient problem. The attention mechanism here only focuses on the feature maps with spatial resolutions equal to or larger than that of the target feature map outputs. Strided convolution with stride $s_ij$ is used to reduce spatial resolution of the feature maps to match target output size and also to match the number of output features map as per the attention configuration. Fig~\ref{fig:artelus_oct3} shows an overview of the partial attention mechanism used, where $E_i$ are the current target feature maps for attention to applying (before reducing the feature map width and height for each block) of each convolution block of the classifier.  Attention weights $a_i$ are learned using attention parameters $W$. Feature-wise attention is applied over the convolved feature maps and summed up using the attention weights. The attended features map is again concatenated with the target feature map to preserve the original information. To reduce aliasing effect we use another convolutional layer to get the attended output $E_{i_attended}$.

\begin{equation}
s_ij= \frac{min(height_{E_{i}}, width_{E_{i}})}{min(height_{E_{j}}, width_{E_{j}})}
\end{equation}

\begin{figure}[h]
  \centering
      \includegraphics[width=3.1in,height=2.6in]{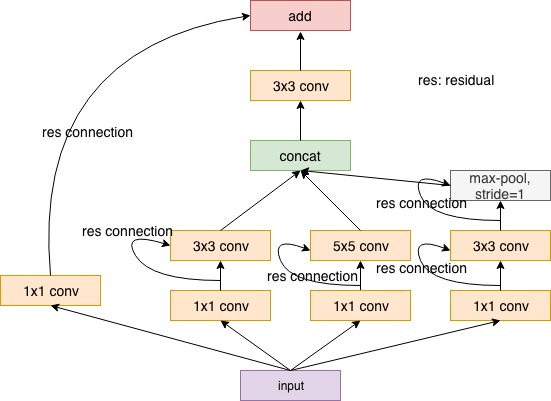}
\caption{Residual Inception}
\label{fig:incept}
\end{figure}

\begin{figure}[h]
  \centering
      \includegraphics[width=3.1in,height=2.9in]{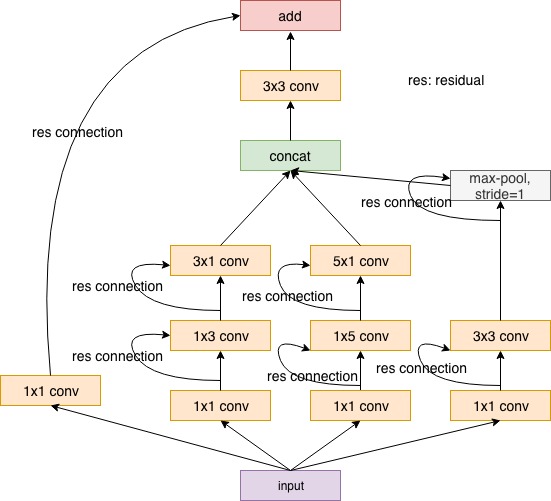}
\caption{Residual Inception with convolution factorization for the classifier node}
\label{fig:incept2}
\end{figure}

\textbf{Dense Residual Inception module} We have used an improved inception block proposed in \cite{haloi} as a baseline of our new inception blocks, where block-based convolution layers, residual and skip connections were used to ensure rich feature representation. We have added a primary residual connection from the clock input to the output as shown in Fig~\ref{fig:incept} to augment the network with block input features and also to solve the vanishing gradient problem for a very deep network. Apart from that, we have used a residual connection in the $3x3$ and $5x5$ convolutional layers of the block; these connections improve the network feature representations capability and also slightly reduces training convergence time. The final concatenated intermediate output features map of the block is convolved with another $3\times3$ convolutional layer to reduce aliasing. A residual connection is added with $1\times1$ convolved input features to make sure that both of the features may have the same number of output filters. Additionally, we have used a convolution factorization version Fig~\ref{fig:incept2} of the same module for the classifier. Convolutional factorization adds efficiency of doing a convolution of a specific spatial dimension with low computational cost.


\section{EXPERIMENTS AND RESULTS}
\textbf{Dataset:} We have used a Mendeley dataset with 4 distinct classes (CNV, DME, DRUSEN, NORMAl) to train the AI system for OCT analysis. Also for the second level transfer learning ChestXray-14 dataset is used to pre-train the
network. ChestXray-14 dataset has around 89K cases with 14 distinct labels. 

\textbf{Metrics:} The network performance was measured using the sensitivity and the specificity. We have also computed precision, recall, and F1-score to further test the performance.

\textbf{Transfer Learning:}
Two-step hierarchical transfer learning is used in this work. In the second step, we have used the variational auto-encoder based transfer learning ~\ref{net_arch} on the training set with the loose label.  In the first step, the weight of the variational auto-encoder is initialized using the pre-trained model on the ChestXray-14 \cite{chestxray14} dataset. It learns diverse and discriminative features of the chest X-ray images which holds pixel based resemblance with OCT images.  Features knowledge of the trained auto-encoder model is transferred to the final classification model trained on the labeled set.

\textbf{Preprocessing:}
For the classifier, $224\times224\times3$ is used as the input size. All input images were resized to $256\times256\times3$ and a randomly cropped patch of size $224\times224\times3$ is used as input. Extensive data augmentation is used such as random flip left/right, up/down and changing the image pixel values randomly using hue, contrast, and saturation. Also, each image is standardized by it's mean and dividing its standard deviation. Apart from that, we have also used random masking of the training images. For each of the training images, we generate 5 to 8 random masks of varied width and height. Random masking proved to be useful for generalization of the deep model.

\begin{figure}[h]
  \centering
      \includegraphics[width=3.1in,height=2.1in]{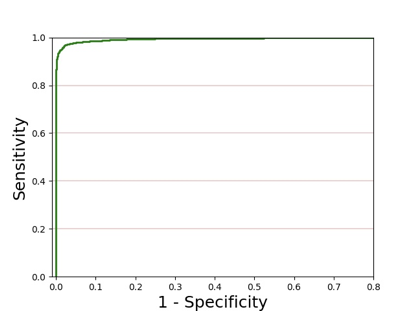}
\caption{ROC plot for OCT}
\label{fig:octfprtpr}
\end{figure}

\begin{figure}[h]
  \centering
      \includegraphics[width=3.5in,height=2.6in]{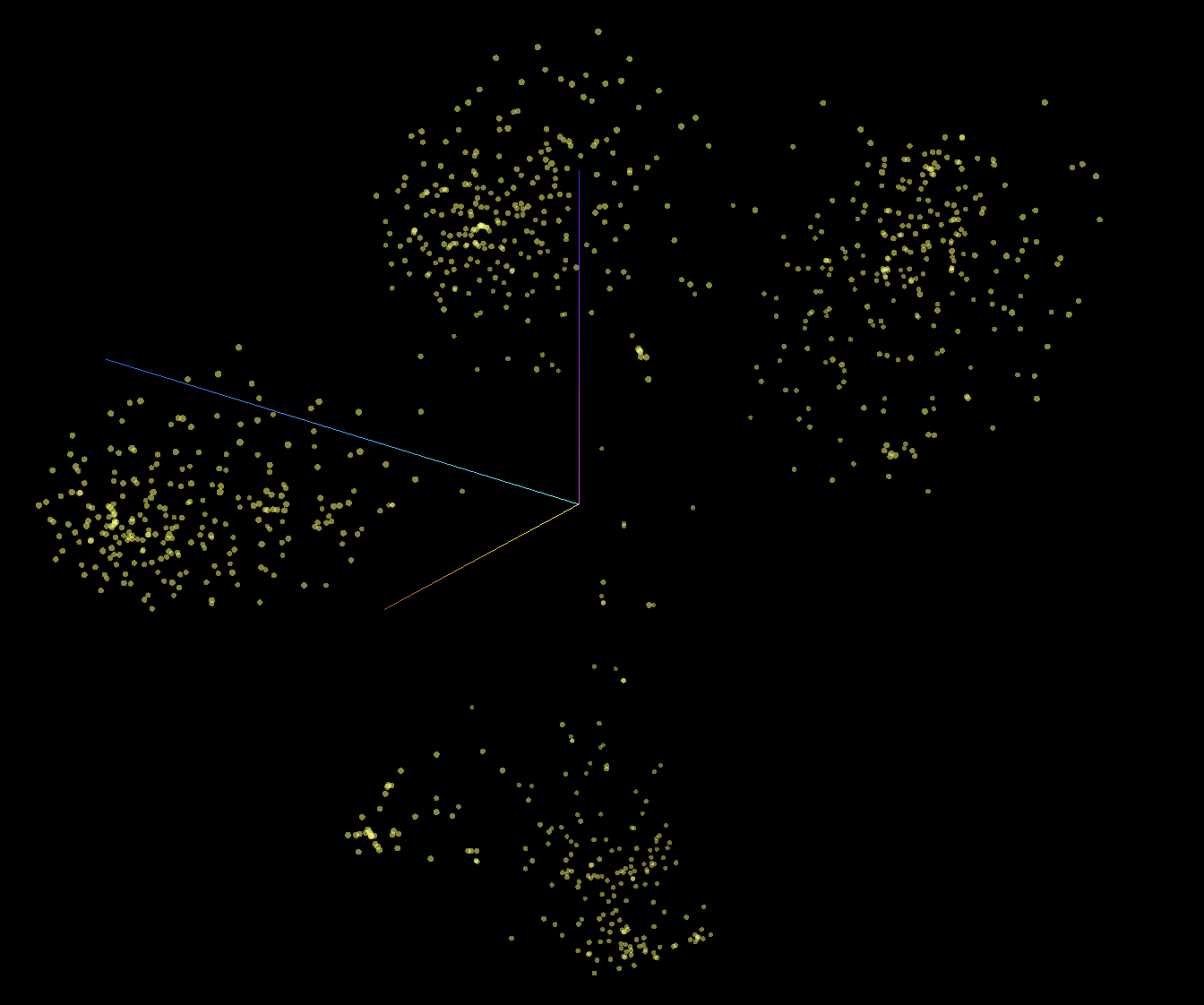}
\caption{T-SNE plot of 1000 test samples using the deep model presented in this work.}
\label{fig:tsnetest}
\end{figure}

\begin{figure}[h]
  \centering
      \includegraphics[width=3.5in,height=4.3in]{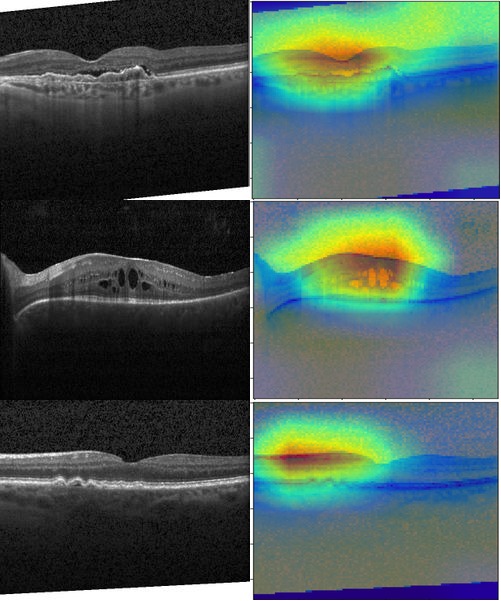}
\caption{Heatmap of the model: Region of interest the model is looking at while making prediction.}
\label{fig:heatmap_oct}
\end{figure}

\begin{table}[h]
\begin{center}
  \begin{tabular}{ | p{3cm} | c |c|c| }
    \hline
    Diseases & Sensitivity (\%) & Specificity (\%) & AUC \\ \hline
    Normal & 100 & 100 & 1.0 \\ 
    abnormal  & 100 & 100 & 1.0 \\
    \hline
  \end{tabular}
\end{center}
\caption{Model performance for binary normal vs abnormal classification on the test set}
\label{table:comp}
\end{table}

\begin{table}[h]
\begin{center}
  \begin{tabular}{ | p{3cm} | c |c|c| }
    \hline
    Diseases & Precision (\%) & Recall (\%) & F1-score \\ \hline
    Normal & 100 & 100 & 1.0 \\ \hline
    CNV & 99 & 100 & 1.0 \\\hline
    DME & 100 & 100 & 1.0 \\\hline
    Drusen & 100 & 100 & 1.0 \\\hline
  \end{tabular}
\end{center}
\caption{Model performance for multi-class classification on the test set}
\label{table:comp_mc}
\end{table}

\textbf{Framework:} We have used TEFLA\cite{tefla}, a python framework developed on the top of TENSORFLOW\cite{tf}, for all experiments described in this work.

\textbf{Training procedure:} Batch normalization\cite{bn} is used to reduce covariate shift and achieve faster convergence of both models. Also, we have used Nesterov momentum optimizer with polynomial learning rate policy. Gradient normalization\cite{gn} was used to stabilize the training process. To improve generalization capability we
have used label smoothing (soft targets). Dropout and batch
sample balancing technique was used to ensure that the network doesn't overfit.
Five models with different
architectures were trained on the same training and
validation set. For the pre-trained ChestXray-14 model used for second level transfer learning, we have also explored weight initializations from the same network trained on the Imagenet \cite{imagenet} dataset.

\textbf{Results analysis:} We have done extensive performance analysis using binary and multi-class classification metrics. Figure~\ref{fig:octfprtpr} shows the receiver operating curve (AUC) for normal vs disease classification. It can be seen that our classifier has very high sensitivity and specificity surpassing human ophthalmologists level performance. 
The trained deep model was able to learn very discriminative features. We have computed T-SNE visualization of 1000 test samples features computed by the learned model. Figure~\ref{fig:tsnetest} shows the T-SNE visualization of the test samples; the features for different classes are widely separated and the model was able to learn proper discrimination among different oct lesions.

Apart from T-SNE analysis, we have also validated our model prediction using heat map analysis. The heat map shows the region of interest the model is looking at while making the prediction. Figure~\ref{fig:heatmap_oct} shows the Artelus AI model heat map on a few sample images. The model was able to learn retinal lesions responsible for sight-threatening diseases.

Table~\ref{table:comp} shows the model performance on the test dataset for normal vs abnormal classification. We have achieved 100\% sensitivity and 100\% specificity for normal vs abnormal task. Table~\ref{table:comp_mc} shows the multi-class performance of the model on the test set. The AI system can screen with expert ophthalmologist level performance. 

\section{Conclusion}
In this work of OCT images, a deep learning
based AIAS system with high sensitivity and specificity for
detecting eye diseases using OCT scans is presented. The proposed AIAS system with sensitivity 100\%  and specificity 100\% on our test dataset will enhance accuracy and response time for OCT-based diagnosis. Temporal consistency of grading across OCT images for a specific operating point is an added efficiency of an automated AIAS system. 
This model could be used for screening and the development of computer-aided diagnosis tools in the future for management of progression of retinal diseases by reducing the burden on doctors and making health care available to all the population.



%

\end{document}